\documentclass[10pt,twocolumn,letterpaper]{article}

\usepackage[pagenumbers]{cvpr}

\usepackage{booktabs}
\usepackage{xcolor}
\usepackage{colortbl}
\usepackage{tabularx}

\newcolumntype{Y}{>{\centering\arraybackslash}X}

\newcommand{\graycell}[1]{\textcolor{black!55}{#1}}

\usepackage{placeins}
\definecolor{cvprblue}{rgb}{0.21,0.49,0.74}
\usepackage[breaklinks,colorlinks,allcolors=cvprblue]{hyperref}

\hypersetup{pdftitle={TAED: Trajectory-Aligned Encoder-Decoder Distillation for Detection Transformers}}

\title{TAED: Trajectory-Aligned Encoder--Decoder Distillation\\for Detection Transformers}

\author{%
  Zilong He\textsuperscript{1}\quad
  Zhangchi Hu\textsuperscript{1}\quad
  Hebei Li\textsuperscript{1}\quad
  Yijun Pan\textsuperscript{1}\quad
  Yifan Zhao\textsuperscript{1}\\
  Xiangyu Yu\textsuperscript{1}\quad
  Yansong Peng\textsuperscript{1}\quad
  Xinghao Wang\textsuperscript{2}\quad
  Dongsheng Jiang\textsuperscript{2}\quad
  Xiaoyan Sun\textsuperscript{1,*}\\[3pt]
  {\normalsize\textsuperscript{1}University of Science and Technology of China}\\
  {\normalsize\textsuperscript{2}Huawei Technologies Ltd}\\[2pt]
  {\small\textsuperscript{*}Corresponding author}%
}
\hypersetup{pdfauthor={Zilong He, Zhangchi Hu, Hebei Li, Yijun Pan, Yifan Zhao, Xiangyu Yu, Yansong Peng, Xinghao Wang, Dongsheng Jiang, Xiaoyan Sun}}

\begin{document}
\maketitle
\begin{abstract}

Distilling localization knowledge into compact detection transformers (DETRs) requires consistent teacher–student query correspondence across decoder layers. However, in DETRs with encoder-based query selection, the teacher and student select top-\(K\) proposals independently. Same-rank queries can originate from different spatial locations and correspond to different objects, making rank-based layer-wise supervision unreliable. To address this problem, we propose Trajectory-Aligned Encoder–Decoder Distillation (TAED), a framework for knowledge transfer between DETRs of different capacities. Encoder feature distillation matches normalized multi-scale features to strengthen student representations and increase the number of teacher–student query pairs from the same encoder locations. Source-indexed query alignment pairs native queries from the same encoder locations and retains pairs associated with the same object for which the teacher provides reliable and more accurate final localization than the student. Trajectory-aligned localization distillation maintains these correspondences and uses teacher proposals and corresponding decoder-layer box predictions to guide student refinement, without introducing additional queries. On COCO, TAED improves DINO-R18 and two-stage Deformable DETR-R18 by 2.5 and 1.7 AP, respectively. Further gains on D-FINE and ECDet demonstrate applicability to recent real-time DETRs at no additional inference cost.
\end{abstract}

\begin{figure}[!t]
\centering
\includegraphics[width=\columnwidth]{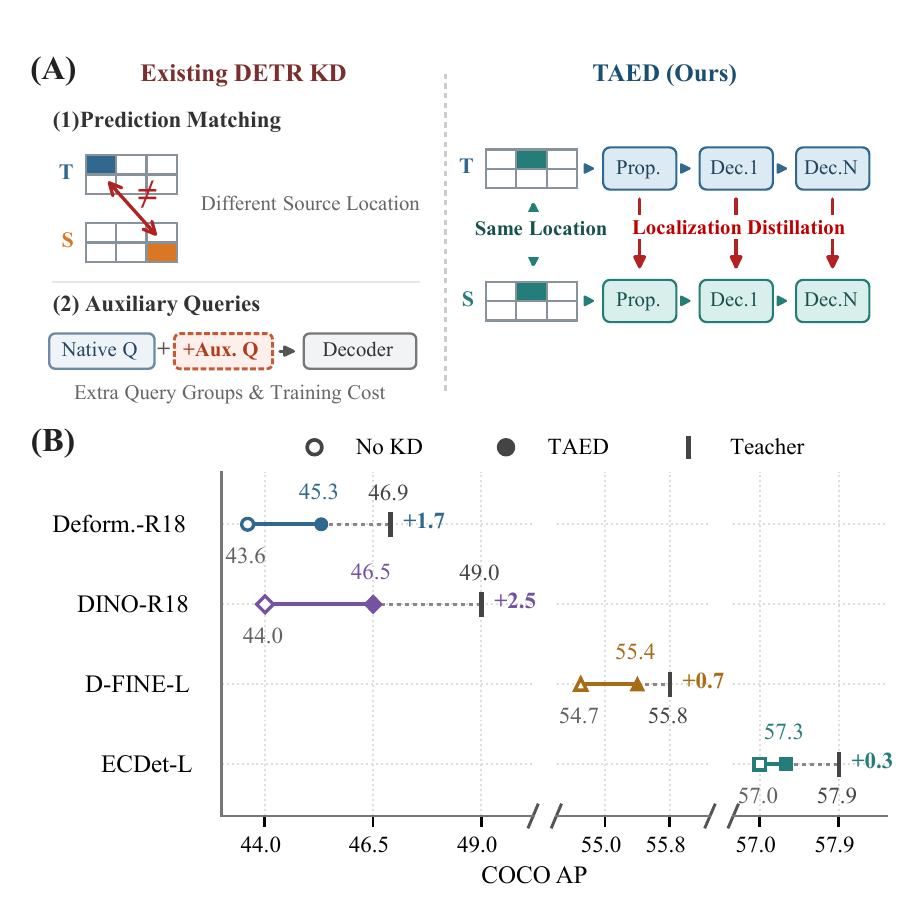}

\caption{\textbf{Query correspondence and cross-family results.} (A) Comparison of query correspondence in existing methods and TAED. (B) AP gains of TAED across DETR families on COCO val2017 without additional inference cost.}

\label{fig:teaser}
\end{figure}

\section{Introduction}
\label{sec:intro}

Object detection is fundamental to autonomous driving and embodied intelligence, where high accuracy, low latency, and computational efficiency are critical~\cite{geiger2012kitti,shridhar2020alfred,RTDETR}. Detection Transformers (DETRs) offer a streamlined end-to-end paradigm, yet practical deployment still demands a tight balance between accuracy and inference cost~\cite{carion2020end,RTDETR}. Scaling up model capacity improves accuracy at the expense of higher latency~\cite{D-FINE}. In contrast, knowledge distillation (KD) transfers supervision from a strong teacher during training, improving the detection accuracy of compact students without increasing their inference cost~\cite{hinton2015distilling,chang2023detrdistill}.

However, effective localization distillation in DETRs hinges on consistent query correspondence between teacher and student. Modern DETRs widely adopt encoder-based query selection, choosing top-$K$ encoder proposals to initialize decoder boxes for iterative refinement~\cite{zhu2020deformable,zhang2022dino,RTDETR,D-FINE}. Because teacher and student select top-$K$ proposals independently, same-rank queries can originate from different spatial locations and correspond to different objects. Consequently, naive rank-based supervision across decoder layers becomes unreliable. Furthermore, coherent layer-wise supervision requires each distilled student query to maintain a fixed teacher pairing throughout all decoder layers.

Prior DETR distillation methods establish query correspondence via output matching or auxiliary query groups (Figure~\ref{fig:teaser}A, left). DETRDistill applies Hungarian matching to output predictions~\cite{chang2023detrdistill}, whereas D$^3$ETR combines adaptive output matching with fixed matching on shared input queries~\cite{ijcai2024p74}. Yet output-level matching fails to ensure that paired native queries originate from the same encoder location. Alternatively, KD-DETR~\cite{wang2024kd} and CLoCKDistill~\cite{lan2026clockdistill} introduce dedicated distillation or ground-truth-conditioned queries, restricting alignment to extra query sets while increasing training cost. Under aligned encoder spatial layouts, pairing native queries directly by shared source location eliminates auxiliary query groups. Nonetheless, naive location-based pairing faces three key obstacles. Independent top-$K$ selection leaves some student queries without a teacher query from the same encoder location. Shared encoder locations still do not guarantee identical target objects. Moreover, even on the same object, the teacher's final prediction may be unreliable or less accurate than the student's.

To overcome these obstacles, we present \textbf{Trajectory-Aligned Encoder--Decoder Distillation (TAED)}, a unified framework that combines encoder feature distillation with localization supervision between native decoder queries (Figure~\ref{fig:teaser}A, right). TAED integrates three complementary components. \emph{Encoder feature distillation} aligns normalized multi-scale features, strengthening student representations and increasing the number of teacher–student query pairs from the same encoder locations. \emph{Source-indexed query alignment} pairs native queries from identical encoder locations and retains only object-consistent pairs where the teacher yields reliable and more accurate final localization than the student. \emph{Trajectory-aligned localization distillation} preserves these verified correspondences across layers, leveraging teacher proposals and decoder box predictions to guide iterative student refinement. Operating without auxiliary query groups, TAED incurs no extra query overhead in training and leaves the student's inference architecture and cost unchanged. Across diverse DETR families on COCO, TAED consistently narrows the accuracy gap toward stronger teachers (Figure~\ref{fig:teaser}B). Specifically, it improves DINO-R18 by 2.5 AP and two-stage Deformable DETR-R18 by 1.7 AP, while further boosting competitive real-time detectors D-FINE-L (DEIM) and ECDet-L by 0.7 and 0.3 AP, respectively.

Our main contributions are summarized as follows:
\begin{itemize}
\item \textbf{Source-Indexed Query Alignment.} We establish object-consistent, quality-filtered correspondence directly between native teacher and student queries from identical encoder locations.
\item \textbf{Trajectory-Aligned Localization Distillation.} We formulate layer-wise localization supervision with fixed teacher–student query correspondence from proposal initialization through decoder refinement, forming a unified KD framework with encoder feature distillation and without auxiliary query groups.
\item \textbf{Broad Applicability Across DETR Families.} Extensive experiments on COCO demonstrate consistent gains across both standard architectures (DINO, two-stage Deformable DETR) and real-time detectors (D-FINE, ECDet) over multiple student capacities.
\end{itemize}

\section{Related Work}
\label{sec:related}

\paragraph{Detection transformers.}
DETR formulates object detection as end-to-end set prediction~\cite{carion2020end}. Deformable DETR uses sparse multi-scale attention, while Conditional DETR introduces query-conditioned spatial cross-attention~\cite{zhu2020deformable,meng2021conditional}. Anchor DETR associates queries with anchor points, and DAB-DETR uses dynamically updated anchor boxes~\cite{wang2022anchor,liu2022dab}. Efficient DETR uses dense prediction priors for query initialization~\cite{yao2021efficient}. Training advances include query denoising in DN-DETR, group-wise one-to-many assignment in Group DETR, and improved denoising and query initialization in DINO~\cite{li2022dn,chen2023group,zhang2022dino}. These developments improve query learning and localization within individual detectors. For real-time detection, RT-DETR and LW-DETR emphasize efficient architectures~\cite{RTDETR,chen2024lw}. D-FINE introduces fine-grained distribution refinement, while DEIM improves matching and training efficiency~\cite{D-FINE,DEIM}. DEIMv2 and EdgeCrafter further explore foundation-model representations and compact ViT backbones~\cite{deimv2,EdgeCrafter}. TAED complements these developments by transferring knowledge between detectors. It exploits encoder-based query selection to establish native-query correspondence between teachers and students with corresponding encoder feature layouts.

\paragraph{Feature distillation for object detection.}
Early methods transfer intermediate features, attention maps, and relationships between representations~\cite{FitNets,Zagoruyko2017AT,relationKD}. Detection-specific approaches combine feature and prediction supervision, with subsequent methods emphasizing object regions and informative instances~\cite{chen2017learning,wang2019distilling,GIbox}. DeFeat separates foreground and background contributions, while FGD combines region-aware feature supervision with global relations~\cite{guo2021defeat,yang2022fgd}. Channel-wise distillation transfers spatial probability maps within feature channels, and MGD reconstructs teacher representations from masked student features~\cite{shu2021cwd,yang2022mgd}. PKD mitigates the influence of feature-magnitude differences through normalized feature matching~\cite{cao2022pkd}. More recently, RT-DETRv4 transfers foundation-model semantics to detector features, while EdgeCrafter learns compact backbones through task-specialized distillation~\cite{RTDETRv4,EdgeCrafter}. In DETRs with encoder-based query selection, encoder feature supervision alone does not guarantee that the teacher and student select the same top-\(K\) query set. Feature similarity alone also does not establish which decoder predictions correspond to the same object. TAED therefore complements normalized encoder feature distillation with explicit native-query pairing for localization supervision across decoder layers.

\paragraph{Query correspondence in DETR distillation.}
Query correspondence is central to prediction-level knowledge transfer between DETRs. Teach-DETR uses teacher-predicted boxes as parallel supervision for student DETRs~\cite{huang2023teach}. DETRDistill combines Hungarian-matched prediction distillation with target-aware features and query-prior assignment~\cite{chang2023detrdistill}. D$^3$ETR uses MixMatcher to combine adaptive output matching with fixed matching based on shared input queries, supervising decoder predictions and attention maps~\cite{ijcai2024p74}. KD-DETR introduces specialized distillation queries, while CLoCKDistill constructs ground-truth-conditioned queries and transfers encoder context~\cite{wang2024kd,lan2026clockdistill}. OD-DETR uses an EMA teacher for online supervision and constructs teacher-derived auxiliary query groups~\cite{wu2024od}. Output matching associates predictions, while shared query inputs provide consistent distillation points; these strategies can coexist within one method. TAED instead pairs independently selected native queries by encoder source index and retains reliable pairs for supervision. Their correspondence remains fixed from initial proposals through decoder layers, enabling layer-wise localization distillation without introducing auxiliary query groups during training.

\section{Method}

\begin{figure*}[t]
\centering
\includegraphics[width=\textwidth]{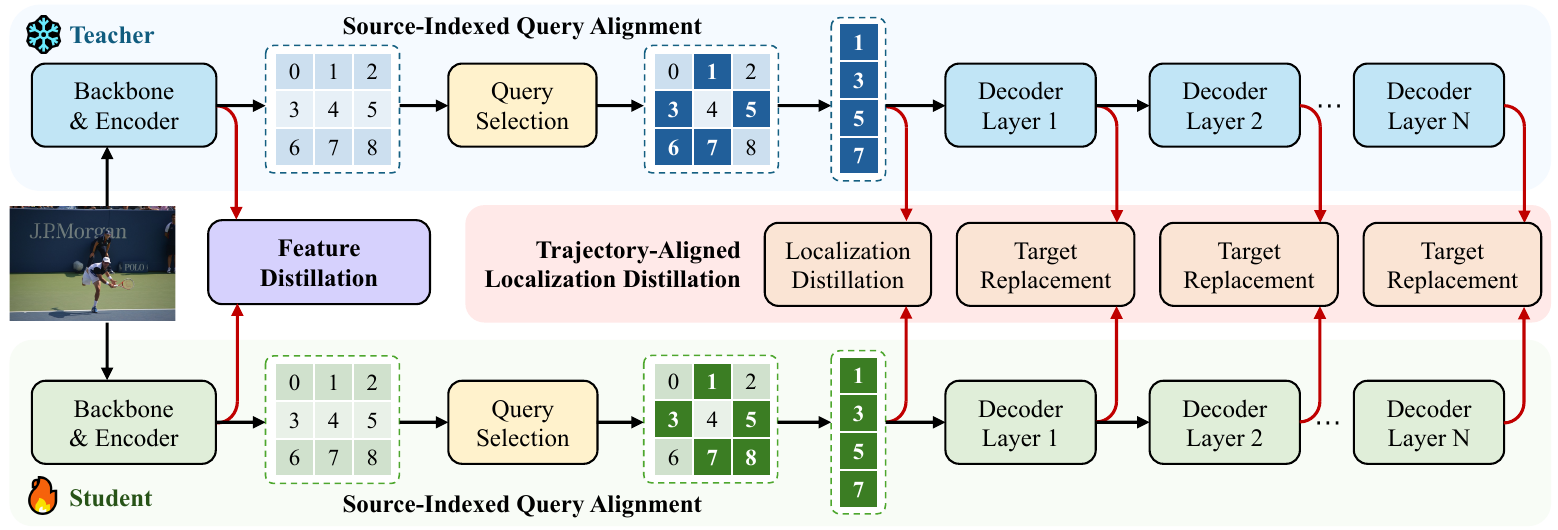}

\caption{\textbf{Overview of TAED.} Encoder feature distillation increases the number of teacher–student query pairs from the same encoder locations, while source-indexed query alignment selects reliable pairs. Trajectory-aligned localization distillation uses fixed query correspondence for proposal supervision and layer-wise localization target replacement.}

\label{fig:overview}
\end{figure*}

We consider knowledge distillation between a frozen teacher $T$ and a trainable student $S$ of different capacities within the same DETR family. As illustrated in Figure~\ref{fig:overview}, TAED combines encoder feature distillation, source-indexed query alignment, and trajectory-aligned localization distillation to transfer representation and localization knowledge without changing the student's inference architecture or computational cost.

\subsection{Preliminaries}
\label{sec:preliminaries}

\paragraph{Query selection and source indices.}
In encoder-based query selection, the detector selects the top-$K$ queries based on encoder confidence scores. Each selected query is associated with an initial proposal, which is progressively refined across decoder layers. Let $a_i^S$ and $a_j^T$ denote the encoder source indices of student query $i$ and teacher query $j$. These indices identify encoder source locations rather than query ranks in the selected top-$K$ sets. The teacher and student use the same indexing convention, so equal source indices identify the same feature-map level and spatial position in both models.

\paragraph{Localization trajectories.}
Let $\mathbf b_{i,n}^S$ and $\mathbf b_{j,n}^T$ denote the student and teacher box predictions at decoder layer $n=1,\ldots,N$, where $N$ is the number of decoder layers. We use $n=0$ to denote their initial proposals. A query's localization trajectory consists of its initial proposal and the box predictions at successive decoder layers.

\subsection{Encoder Feature Distillation}
\label{sec:encoder_kd}

We use encoder feature distillation to strengthen student representations and increase the number of query pairs with the same encoder source index between the teacher's and student's top-$K$ sets. Following PKD~\cite{cao2022pkd}, we match normalized multi-scale encoder features between the teacher and student before query selection.

When channel dimensions differ, a training-only $1\times1$ convolution projects student features to the teacher's channel dimension. Let $F_l^S$ and $F_l^T$ denote the resulting student features and the corresponding teacher features at level $l$. We denote channel-wise standardization over the batch and spatial dimensions by $\mathcal N(\cdot)$. This standardization reduces the influence of feature-magnitude differences between the teacher and student. The encoder loss is
\begin{equation}
\mathcal L_{\mathrm{enc}}
=\sum_l
\operatorname{MSE}\!\left(
\mathcal N(F_l^S),\mathcal N(F_l^T)
\right),
\label{eq:encoder_kd}
\end{equation}
where $l$ indexes the encoder feature levels used for distillation,
and MSE denotes the mean squared error over all feature elements at each level.

Encoder feature distillation supervises the full feature maps, whereas localization distillation supervises box predictions from selected native query pairs.

\subsection{Source-Indexed Query Alignment}
\label{sec:query_alignment}

TAED combines source alignment with a quality gate to establish reliable correspondence between native queries in the teacher's and student's top-\(K\) sets.

\paragraph{Source alignment.}
Each pair must satisfy $a_i^S=a_j^T$. Student queries used for distillation are selected from the detector's positive set $\mathcal P^S$. For D-FINE-based detectors, we use the global optimal (GO) set, obtained by taking the union of positive queries from Hungarian matching across decoder and auxiliary prediction heads~\cite{D-FINE}.

\paragraph{Quality gate.}
The quality gate retains query pairs associated with the same ground-truth (GT) object for which the teacher provides reliable and more accurate final localization than the student. Let $g_i^S$ denote the GT index assigned to student query $i$, and $g_j^T$ the GT index associated with teacher query $j$ by maximum final-layer IoU. The object-consistent candidate set is
\begin{equation}
\mathcal C=\left\{(i,j)\;\middle|\;
i\in\mathcal P^S,\ a_i^S=a_j^T,\ g_i^S=g_j^T\right\}.
\label{eq:object_pairs}
\end{equation}
For each pair in $\mathcal C$, let $u_i^S$ and $u_j^T$ denote the IoUs of the student and teacher final-layer box predictions with the same GT box. The retained decoder supervision set is
\begin{equation}
\mathcal R=\left\{(i,j)\in\mathcal C
\;\middle|\; u_j^T\geq\tau,\ u_j^T-u_i^S\geq\delta\right\},
\label{eq:decoder_pairs}
\end{equation}
where $\tau$ denotes the teacher IoU threshold and $\delta$ the teacher--student IoU margin. The absolute threshold excludes low-quality teacher boxes even when their IoU exceeds the student's, while the margin requires a sufficient localization advantage. The selected query pairs remain fixed across decoder layers.

\subsection{Trajectory-Aligned Localization Distillation}
\label{sec:trajectory_kd}

TAED transfers localization knowledge from initial proposals through decoder refinement using fixed teacher--student query pairs. Proposal supervision reduces localization discrepancies between paired initial boxes, while decoder target replacement guides subsequent refinement using predictions from corresponding teacher layers. For each query pair, let $\mathbf b_n^S$ and $\mathbf b_n^T$ denote its layer-$n$ box predictions, and $\mathbf b^{\mathrm{gt}}$ the corresponding GT box.

\paragraph{Proposal localization distillation.}
For candidate pairs in $\mathcal C$, we supervise student initial proposals using the paired teacher proposals. The proposal distillation loss is
\begin{equation}
\mathcal L_{\mathrm{prop}}
=\lambda_{\mathrm{bbox}}\mathcal L_{\mathrm{bbox}}
+\lambda_{\mathrm{giou}}\mathcal L_{\mathrm{giou}},
\label{eq:proposal_kd}
\end{equation}
where $\mathcal L_{\mathrm{bbox}}$ and $\mathcal L_{\mathrm{giou}}$ denote the L1 coordinate regression and GIoU losses~\cite{rezatofighi2019giou} between paired student and teacher proposals, respectively. The coefficients $\lambda_{\mathrm{bbox}}$ and $\lambda_{\mathrm{giou}}$ balance the two terms. Both terms are weighted according to the teacher proposal’s IoU with the GT and its IoU advantage over the student proposal. This loss is added to the native training objective to reduce localization discrepancies between paired proposals.

\paragraph{Layer-wise localization target replacement.}
Let $\mathbf t_n$ denote the localization target for a student positive query at decoder layer $n$. It is defined as
\begin{equation}
\mathbf t_n=
\begin{cases}
\mathbf b_n^T,&\text{if paired in }\mathcal R,\\
\mathbf b^{\mathrm{gt}},&\text{otherwise}.
\end{cases}
\label{eq:routed_box}
\end{equation}
Each selected student query is supervised by predictions from the corresponding layers of the same teacher query. These predictions replace the GT targets in the student's native L1 and GIoU losses, rather than introducing an additional decoder box-distillation loss.

\paragraph{Adaptation to D-FINE-based decoders.}
D-FINE-based detectors additionally use fine-grained localization (FGL) and decoupled distillation focal (DDF) losses~\cite{D-FINE}. For selected queries, the teacher's final-layer box replaces the GT box when constructing FGL targets, which are reused across auxiliary decoder layers. This differs from the layer-matched box targets used for L1/GIoU supervision.

Selected student queries are excluded from the positive DDF term without being reassigned to the negative branch. DDF supervision for the remaining queries is unchanged.

\subsection{Overall Training Objective}
\label{sec:objective}

The overall training objective is
\begin{equation}
\mathcal L=\mathcal L_{\mathrm{det}}
+\lambda_{\mathrm{enc}}\mathcal L_{\mathrm{enc}}
+\mathcal L_{\mathrm{prop}},
\label{eq:overall_objective}
\end{equation}
where $\mathcal L_{\mathrm{det}}$ denotes the student's detection loss with the target replacement and detector-specific adaptations described above, and $\lambda_{\mathrm{enc}}$ is the encoder distillation weight.

The teacher and feature projection layers are used only during training, leaving the student's inference architecture and computational cost unchanged.

\begin{table*}[t]
\centering
\caption{\textbf{Distillation results across DETR families on COCO val2017.} We compare TAED with existing distillation methods on two-stage Deformable DETR, DINO, and D-FINE. D-FINE-L follows the DEIM training recipe. Gray text denotes teachers, and shaded rows highlight TAED. $\dagger$ denotes teachers pretrained on Objects365 and fine-tuned on COCO.}
\normalsize
\setlength{\tabcolsep}{3.5pt}
\renewcommand{\arraystretch}{1.08}
\begin{tabular}{@{}>{\columncolor{white}[0pt][\tabcolsep]}llcccccccc>{\columncolor{white}[\tabcolsep][0pt]}c@{}}
\toprule
\textbf{Model} & \textbf{Teacher} & \textbf{Params} & \textbf{GFLOPs} & \textbf{Latency (ms)} & \textbf{AP} & {\bfseries\boldmath AP$_{50}$} & {\bfseries\boldmath AP$_{75}$} & {\bfseries\boldmath AP$_S$} & {\bfseries\boldmath AP$_M$} & {\bfseries\boldmath AP$_L$} \\
\midrule
\graycell{Deformable-R50 (2-stage)}  & \graycell{--} & \graycell{41M} & \graycell{--} & \graycell{--} & \graycell{46.9} & \graycell{--} & \graycell{--} &\graycell{29.6} & \graycell{50.1} & \graycell{61.6} \\
Deformable-R18 (2-stage)  & -- & 29M & -- & -- & 43.6 & 62.3 & 47.4 & 26.7 & 45.9 & 58.6 \\
\rowcolor{gray!15}
\textbf{+ TAED}  & Deformable-R50 & 29M  & -- & -- & \textbf{45.3} & \textbf{63.6} & \textbf{49.1} & \textbf{27.2} & \textbf{48.0} & \textbf{60.5} \\
\midrule

\graycell{DINO-R50}  & \graycell{--} & \graycell{47M} & \graycell{--} & \graycell{--} & \graycell{49.0} & \graycell{66.4} & \graycell{53.3} & \graycell{31.4} & \graycell{52.2} & \graycell{64.0} \\
DINO-R18  & -- & 30M & -- & -- & 44.0 & 61.2 & 48.1 & 27.4 & 46.9 & 56.9 \\
+ KD-DETR \cite{wang2024kd} & DINO-R50 & 30M & -- & -- & 46.1 & 63.1 & 50.1 & 29.2 & 48.7 & 61.1 \\
+ CLoCKDistill \cite{lan2026clockdistill} & DINO-R50 & 31M & -- & -- & 46.4 & 63.3 & 50.4 & 28.6 & 48.7 & \textbf{61.4} \\
\rowcolor{gray!15}
\textbf{+ TAED} & DINO-R50 & 30M & -- & -- & \textbf{46.5} & \textbf{63.5} & \textbf{50.7} & \textbf{30.2} & \textbf{49.2} & 60.3 \\
\midrule
\graycell{D-FINE-X$^\dagger$} &\graycell{--} &\graycell{62M} & \graycell{202} & \graycell{12.89} &\graycell{59.3} &\graycell{76.8} &\graycell{64.6} &\graycell{42.3} & \graycell{64.2} & \graycell{76.4} \\
\graycell{D-FINE-X} & \graycell{--} & \graycell{62M} & \graycell{202} & \graycell{12.89} & \graycell{55.8} & \graycell{73.7} & \graycell{60.2} & \graycell{37.3} & \graycell{60.5} & \graycell{73.5} \\
D-FINE-L (DEIM) & -- & 31M & 91 & 8.07 & 54.7 & 72.4 & 59.4 & 36.9 & 59.6 & 71.8 \\
+ KD-DETR \cite{wang2024kd} & D-FINE-X & 31M & 91 & 8.07 & 54.9 & 72.6 & 59.6 & 36.6 & 59.6 & 72.3 \\
+ DETRDistill \cite{chang2023detrdistill} & D-FINE-X & 31M & 91 & 8.07 & 54.9 & 72.5 & 59.6 & 38.1 & 59.3 & 72.2 \\
+ D$^3$ETR \cite{ijcai2024p74} & D-FINE-X & 31M & 91 & 8.07 & 54.8 & 72.8 & 59.6 & 36.8 & 59.6 & 72.2 \\
+ RT-DETRv4 \cite{RTDETRv4} & DINOv3 & 31M & 91 & 8.07 & 55.4 & 73.0 & 60.3 & 37.1 & 60.1 & 72.9\\
\rowcolor{gray!15}
\textbf{+ TAED} & D-FINE-X & 31M & 91 & 8.07 & 55.4 & 73.0 & 60.1 & 37.4 & 60.3 & 72.7 \\
\rowcolor{gray!15}
\textbf{+ TAED} & D-FINE-X$^\dagger$ & 31M & 91 & 8.07 & \textbf{56.2} & \textbf{73.8} & \textbf{61.1} & \textbf{38.9} & \textbf{60.8} & \textbf{73.3} \\
\bottomrule
\end{tabular}
\label{tab:transfer}
\end{table*}

\section{Experiments}
\label{sec:experiments}

\subsection{Experimental Setup}

\paragraph{Dataset and metrics.}
We use COCO for student training and evaluation~\cite{lin2014microsoft}, and Objects365 as an additional source of teacher pre-training data~\cite{shao2019objects365}. COCO train2017 and val2017 contain approximately 118,000 training images and 5,000 validation images, respectively. We report the standard COCO AP averaged over IoU thresholds from 0.50 to 0.95, together with AP$_{50}$, AP$_{75}$, and AP for small, medium, and large objects. Model parameters, GFLOPs, and latency are taken from the corresponding papers.

\paragraph{Teacher--student settings.}
We evaluate TAED across four DETR families through within-family teacher--student distillation: two-stage Deformable DETR, DINO, D-FINE, and ECDet. Deformable DETR and DINO use ResNet-50 teachers and ResNet-18 students~\cite{he2016resnet}. D-FINE-L follows the DEIM training recipe~\cite{DEIM}. We consider two D-FINE-X teachers: one trained on COCO, and the other pretrained on Objects365 and subsequently fine-tuned on COCO. The latter is marked with $\dagger$ in the tables. ECDet-S/M/L are distilled using a shared ECDet-X teacher~\cite{EdgeCrafter}.

\paragraph{Training protocols.}
Students follow the optimization schedules, data augmentation, and input settings of their respective detector training recipes. For D-FINE, query selection retains $K=300$ native queries, and the decoder quality gate uses a teacher IoU threshold of $\tau=0.65$ and a teacher--student IoU margin of $\delta=0.01$. Component and alignment ablations use the D-FINE-X$^\dagger$ teacher and D-FINE-L student trained with the DEIM recipe.

\begin{table*}[t]
\centering
\caption{\textbf{Accuracy--efficiency results on COCO val2017.} TAED is evaluated on ECDet-S/M/L using ECDet-X as the teacher. Other real-time detectors serve as references. Gray text denotes the teacher, and shaded rows highlight TAED.}
\small
\setlength{\tabcolsep}{10pt}
\renewcommand{\arraystretch}{1.08}
\begin{tabular}{@{}>{\columncolor{white}[0pt][\tabcolsep]}lcccccccc>{\columncolor{white}[\tabcolsep][0pt]}c@{}}
\toprule
\textbf{Model} & \textbf{Params} & \textbf{GFLOPs} & \textbf{Latency (ms)} & \textbf{AP} & {\bfseries\boldmath AP$_{50}$} & {\bfseries\boldmath AP$_{75}$} & {\bfseries\boldmath AP$_S$} & {\bfseries\boldmath AP$_M$} & {\bfseries\boldmath AP$_L$} \\
\midrule
\graycell{ECDet-X} & \graycell{49M} & \graycell{151} & \graycell{12.70} & \graycell{57.9} & \graycell{76.0} & \graycell{62.9} & \graycell{38.7} & \graycell{63.4} & \graycell{76.1} \\
\midrule
YOLOv10-S~\cite{wang2024yolov10} & 7M & 22 & 2.52 & 46.3 & 63.0 & 50.4 & 26.8 & 51.0 & 63.8 \\
RT-DETRv4-S~\cite{RTDETRv4} & 10M & 25 & 3.66 & 49.7 & 66.8 & 54.1 & 30.2 & 53.6 & 66.9 \\
DEIMv2-S~\cite{deimv2} & 10M & 26 & 5.78 & 50.9 & 68.3 & 55.1 & 31.4 & 55.3 & 70.3 \\
ECDet-S & 10M & 26 & 5.41 & 51.7 & 69.4 & 55.8 & \textbf{32.3} & 56.4 & 70.5 \\
\rowcolor{gray!15}
ECDet-S \textbf{+ TAED} & 10M & 26 & 5.41 & \textbf{52.3} & \textbf{70.1} & \textbf{56.4} & 32.0 & \textbf{57.2} & \textbf{71.7} \\
\midrule
YOLOv10-M~\cite{wang2024yolov10} & 15M & 59 & 4.74 & 51.1 & 68.1 & 55.8 & 33.8 & 56.5 & 67.0 \\
RT-DETRv4-M~\cite{RTDETRv4} & 19M & 57 & 5.91 & 53.5 & 71.1 & 58.1 & 34.9 & 57.7 & 72.1 \\
DEIMv2-M~\cite{deimv2} & 18M & 52 & 8.80 & 53.0 & 70.2 & 57.6 & 34.2 & 57.4 & 71.5 \\
ECDet-M & 18M & 53 & 7.98 & 54.3 & 72.2 & 58.7 & \textbf{35.9} & 59.1 & 72.7 \\
\rowcolor{gray!15}
ECDet-M \textbf{+ TAED} & 18M & 53 & 7.98 & \textbf{55.1} & \textbf{73.1} & \textbf{59.6} & 35.5 & \textbf{60.3} & \textbf{74.1} \\
\midrule
YOLOv10-L~\cite{wang2024yolov10} & 24M & 120 & 7.38 & 53.2 & 70.1 & 58.1 & 35.8 & 58.5 & 69.4 \\
RT-DETRv4-L~\cite{RTDETRv4} & 31M & 91 & 8.07 & 55.4 & 73.0 & 60.3 & 37.1 & 60.1 & 72.9 \\
DEIMv2-L~\cite{deimv2} & 32M & 96 & 10.47 & 56.0 & 73.4 & 60.9 & 37.5 & 60.8 & 75.2 \\
ECDet-L & 31M & 101 & 10.49 & 57.0 & 75.1 & 61.7 & 38.7 & 62.5 & 75.0 \\
\rowcolor{gray!15}
ECDet-L \textbf{+ TAED} & 31M & 101 & 10.49 & \textbf{57.3} & \textbf{75.4} & \textbf{62.1} & \textbf{39.3} & \textbf{63.1} & \textbf{75.3} \\
\bottomrule
\end{tabular}
\label{tab:main}
\end{table*}

\subsection{Results across DETR Families}

\paragraph{Effectiveness across DETR families.}
Table~\ref{tab:transfer} evaluates TAED on two-stage Deformable DETR, DINO, and D-FINE. TAED improves Deformable DETR-R18 and DINO-R18 by 1.7 and 2.5 AP, respectively. For D-FINE-L trained with the DEIM recipe, TAED raises AP from the reported 54.7 baseline to 55.4 using a COCO-trained D-FINE-X teacher. These results show that its benefits extend from lower-accuracy students to an already accurate real-time detector. AP$_{75}$ also improves for all three students, indicating that the gains persist at an IoU threshold of 0.75. Improvements in AP$_S$, AP$_M$, and AP$_L$ across all three students further show that the benefits are not confined to a particular object-size range. Together, these results support applicability across DETR families with encoder-based query selection and corresponding encoder feature layouts.

\begin{table}[t]
\centering
\caption{\textbf{Component ablations of TAED.} Results are for D-FINE-X$^\dagger$ $\rightarrow$ D-FINE-L (DEIM). Components are added cumulatively; the last row removes encoder feature distillation from full TAED while keeping all other components unchanged.}
\small
\setlength{\tabcolsep}{8pt}
\renewcommand{\arraystretch}{1.12}
\begin{tabular*}{0.94\columnwidth}{@{\extracolsep{\fill}}lc@{}}
\toprule
\textbf{Configuration} & \textbf{AP} \\
\midrule
D-FINE-L (DEIM) baseline & 54.70 \\
\quad + Encoder feature distillation
    & 55.78 \\
\quad + Layer-wise target replacement
    & 56.12 \\
\quad + Proposal localization distillation
    & 56.16 \\
\quad + DDF conflict masking (\textbf{Full TAED})
    & \textbf{56.23} \\
\addlinespace[2pt]
\midrule
\multicolumn{2}{@{}l@{}}{\textit{Removal from full TAED}} \\
w/o Encoder feature distillation & 54.88 \\
\bottomrule
\end{tabular*}
\label{tab:overall_components}
\end{table}

\paragraph{Comparison with distillation methods.}
On DINO-R18, TAED achieves 46.5 AP, compared with the reported 46.1 AP of KD-DETR and 46.4 AP of CLoCKDistill~\cite{wang2024kd,lan2026clockdistill}. TAED also obtains higher AP$_{75}$ than both methods, while CLoCKDistill retains higher AP$_L$. Thus, native query correspondence provides competitive overall and high-IoU detection performance without auxiliary distillation queries.

We implement and evaluate KD-DETR, DETRDistill, and D$^3$ETR~\cite{wang2024kd,chang2023detrdistill,ijcai2024p74} on D-FINE-L using a COCO-trained D-FINE-X teacher and the DEIM training recipe. These methods achieve 54.8--54.9 AP, compared with 55.4 AP for TAED. TAED also matches the overall AP of RT-DETRv4~\cite{RTDETRv4}, which uses DSI/GAM feature distillation from DINOv3~\cite{simeoni2025dinov3}. This comparison supports transferring encoder representations and decoder localization knowledge from a detector teacher as an effective alternative to foundation-model feature supervision in this setting.

\paragraph{Benefit from a stronger teacher.}
Using the Objects365-pretrained D-FINE-X$^\dagger$ teacher further raises D-FINE-L to 56.2 AP, improving on the COCO-teacher result by 0.8 AP. This additional gain shows that TAED can benefit from stronger teacher supervision while retaining the same student architecture, extending its effectiveness to a teacher with additional pre-training data.

\subsection{Accuracy--Efficiency Trade-off}

Table~\ref{tab:main} demonstrates the effectiveness of TAED on ECDet, a high-accuracy real-time DETR family~\cite{EdgeCrafter}. Using ECDet-X as the teacher, TAED improves ECDet-S/M/L by 0.6/0.8/0.3 AP, respectively, while preserving their parameters and GFLOPs. These gains demonstrate improved accuracy at fixed student inference cost across different model capacities. AP$_{75}$ also improves across all three student scales, showing that the gains extend to evaluation at a stricter IoU threshold.

The medium student provides a useful example. ECDet-M with TAED reaches 55.1 AP, approaching the 55.4 AP of RT-DETRv4-L while using 18M rather than 31M parameters and fewer GFLOPs. TAED narrows this accuracy gap from 1.1 to 0.3 AP without increasing student computation.

The benefit also extends to the high-accuracy ECDet-L student, which improves from 57.0 to 57.3 AP. This gain is achieved with an initial teacher--student gap of only 0.9 AP, showing that TAED remains beneficial even when the student is already close to its teacher in AP.

\subsection{Ablation Studies}

\paragraph{Contribution of distillation components.}
Table~\ref{tab:overall_components} first adds encoder feature distillation, followed by the components of trajectory-aligned localization distillation: layer-wise target replacement, proposal supervision, and DDF masking. Target replacement uses both source alignment and the quality gate to select supervision pairs. The localization components improve the encoder-distilled student's AP from 55.78 to 56.23.

\paragraph{Encoder--decoder complementarity.}
Table~\ref{tab:overall_components} shows that localization distillation provides an additional 0.45 AP when combined with encoder feature distillation, compared with 0.18 AP when used alone. The combined framework outperforms either form of distillation alone, supporting the complementary roles of encoder feature and localization supervision in improving student detection accuracy.

\paragraph{Source alignment and quality gate.}
Source-indexed query alignment selects supervision pairs rather than introducing a separate distillation loss. Table~\ref{tab:alignment_ablation} therefore evaluates its two mechanisms within full TAED, with the remaining components unchanged. Replacing source alignment with Hungarian matching reduces AP by 0.30, supporting encoder source indices as a useful constraint for localization correspondence. Removing the quality gate reduces AP by 0.17, showing an additional benefit from selecting reliable teacher supervision.

\begin{table}[t]
\centering
\caption{\textbf{Source alignment and quality-gate ablations.} Results are for D-FINE-X$^\dagger$ $\rightarrow$ D-FINE-L (DEIM). Without source alignment, query pairs are formed by Hungarian matching.}
\small
\setlength{\tabcolsep}{2.75pt}
\begin{tabular*}{\columnwidth}{@{\extracolsep{\fill}}lccc@{}}
\toprule
\textbf{Variant} & \textbf{Source-index align.} & \textbf{Quality gate} & \textbf{AP} \\
\midrule
w/o Quality gate & $\checkmark$ & -- & 56.06 \\
w/o Source alignment & -- & $\checkmark$ & 55.93 \\
\textbf{Full TAED} & $\checkmark$ & $\checkmark$ & \textbf{56.23} \\
\bottomrule
\end{tabular*}
\label{tab:alignment_ablation}
\end{table}

\begin{table}[t]
\centering
\caption{\textbf{Source consistency across pairing strategies.} D-FINE-X$^\dagger$ $\rightarrow$ D-FINE-L (DEIM), evaluated on 5,000 COCO val2017 images. Values are percentages of all student positive queries paired with teacher queries sharing the same encoder source index.}

\label{tab:source_pairing}
\small
\setlength{\tabcolsep}{5pt}
\renewcommand{\arraystretch}{1.10}
\begin{tabular*}{0.98\columnwidth}{@{\extracolsep{\fill}}lcc@{}}
\toprule
\textbf{Pairing strategy} & \textbf{w/o Encoder KD} & \textbf{Encoder KD} \\
\midrule
Rank-based & 10.48 & 17.49 \\
Hungarian & 42.31 & 63.02 \\
\textbf{Source alignment} & \textbf{71.45} & \textbf{89.49} \\
\bottomrule
\end{tabular*}
\end{table}

\begin{table}[t]
\centering
\caption{\textbf{Query-pair selection statistics.} The D-FINE-X$^\dagger$ teacher and D-FINE-L (DEIM) students each select their top-300 native queries. Mean query pairs per image are reported after each step under different encoder feature distillation settings, measured offline using fixed models on COCO val2017 (5,000 images).}
\label{tab:pair_funnel}
\small
\setlength{\tabcolsep}{3pt}
\renewcommand{\arraystretch}{1.10}
\begin{tabular*}{0.98\columnwidth}{@{\extracolsep{\fill}}lccc@{}}
\toprule
\textbf{Query pairs} & \textbf{w/o Encoder KD} & \textbf{RT-DETRv4} & \textbf{Ours} \\
\midrule
Same source & 129.96 & 149.78 & \textbf{174.17} \\
\quad + Student GO & 6.25 & 6.64 & \textbf{7.83} \\
\quad + Same GT & 5.89 & 6.25 & \textbf{7.49} \\
\quad + Quality gate & 1.77 & 1.76 & \textbf{2.30} \\
\bottomrule
\end{tabular*}
\end{table}

\begin{figure*}[t]
\centering
\includegraphics[width=0.95\textwidth]{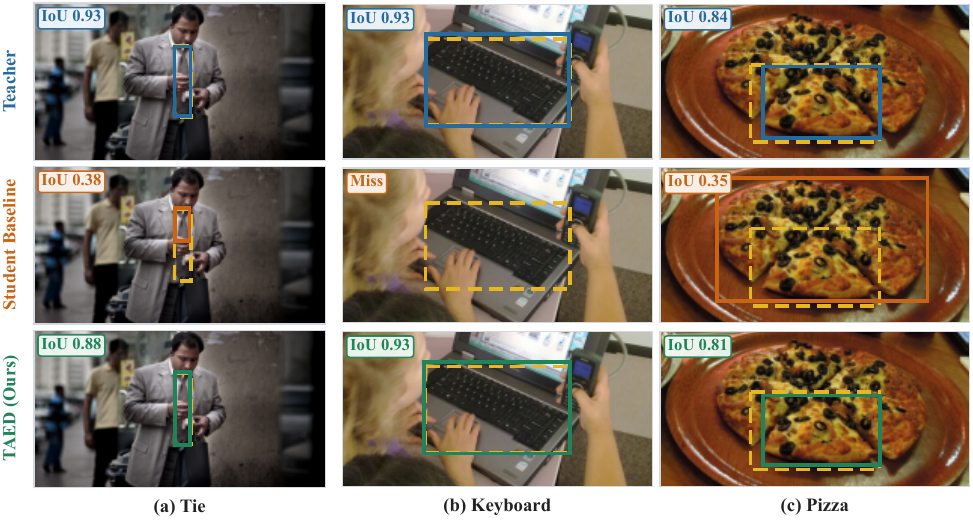}

\caption{\textbf{Qualitative comparison.} Predictions are shown for the D-FINE-X$^\dagger$ teacher, D-FINE-L (DEIM) baseline, and TAED-distilled student. Dashed yellow boxes denote GT; solid boxes denote predictions. IoU is computed against the highlighted GT box.}

\label{fig:qualitative}
\vspace{-4pt}
\end{figure*}

\subsection{Analysis}

\paragraph{Source consistency of query pairing.}
Table~\ref{tab:source_pairing} compares pairing strategies for fixed D-FINE-L models with and without encoder feature distillation. D-FINE-X$^\dagger$ serves as the teacher. Each model selects its top-300 native queries, and percentages are computed over all student positive queries. Rank-based pairing matches queries at the same selection rank. Hungarian matching pairs teacher and student queries based on their final-layer predictions. Source alignment pairs native queries with the same encoder source index. The source-alignment rate is below 100\% because some student positive queries lack a teacher query from the same encoder location. Encoder feature distillation improves same-source pairing across all three strategies, providing more same-source pairs for localization supervision. Source alignment achieves the highest same-source pairing rate both with and without encoder feature distillation. With encoder distillation, it reaches 89.49\%, compared with 63.02\% for Hungarian matching. Table~\ref{tab:alignment_ablation} further shows higher AP with source alignment than with Hungarian matching, supporting the use of encoder source indices to establish query pairs for localization distillation.

\begin{figure}[t]
\centering
\includegraphics[width=0.95\linewidth]{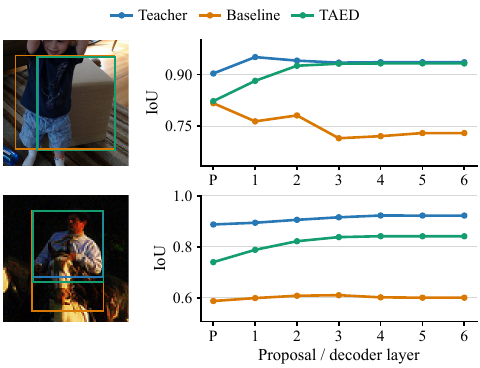}
\caption{\textbf{Final boxes and IoU trajectories.} Two cases showing TAED gains over the baseline are selected from 2,000 COCO val2017 images. On the x-axis, P denotes the initial encoder proposal; 1--6 denote decoder layers.}
\label{fig:trajectory_examples}
\vspace{-14pt}
\end{figure}

\paragraph{Query selection and supervision quality.}
Table~\ref{tab:pair_funnel} reports the mean number of query pairs per image after each step of source-indexed query alignment. For our encoder-distilled student, object-consistency filtering reduces the count only slightly, from 7.83 to 7.49 pairs per image. The quality gate further retains 2.30 pairs whose teacher predictions provide reliable and more accurate final localization than the student. This exceeds the 1.77 pairs retained without encoder feature distillation and 1.76 with RT-DETRv4. These results support complementary roles for the two components: encoder feature distillation increases the availability of same-source pairs, while the quality gate selects reliable teacher predictions that provide more accurate final localization than the student's.

\paragraph{Qualitative results.}
Figure~\ref{fig:qualitative} compares predictions from the teacher, baseline student, and TAED-distilled student. The baseline boxes for the tie and pizza include substantial surrounding background and do not closely fit the target objects. TAED produces boxes with better alignment to the GT, increasing IoU from 0.38 to 0.88 and from 0.35 to 0.81, respectively. These improvements reduce the localization gap to the teacher and provide an instance-level view of the quantitative gains from teacher supervision. The keyboard example additionally illustrates recovery of a detection missed by the baseline.

\paragraph{Localization along fixed query trajectories.}
Figure~\ref{fig:trajectory_examples} visualizes localization from initial proposals through decoder refinement for two selected cases. In both examples, the TAED student narrows the localization gap to the teacher during refinement and achieves higher final IoU than the baseline. The TAED student also improves on its own initial proposals, while the baseline either loses IoU or shows little net improvement during refinement. These examples illustrate why proposal quality alone is insufficient to characterize the full localization process. Fixed teacher--student query correspondence allows predictions from corresponding teacher layers to guide student refinement toward more accurate final boxes for the paired native student queries.

\section{Conclusion}

We presented TAED, an encoder--decoder distillation framework for DETRs with encoder-based query selection. Source-indexed query alignment and quality filtering establish reliable teacher--student query pairs, whose correspondence remains fixed from proposal initialization through decoder refinement. Encoder feature distillation provides more query pairs from the same encoder locations, enabling knowledge transfer through native queries without auxiliary query groups. Experiments on COCO demonstrate accuracy improvements across ECDet, D-FINE, DINO, and two-stage Deformable DETR without increasing student inference cost. The framework requires corresponding encoder feature layouts; extending native-query alignment to detectors with incompatible feature grids remains a direction for future work.

\section*{LLM Usage Statement}
We only exploit the Large Language Model (LLM), i.e., GPT-6, for polishing to improve the presentation quality. No LLMs contributes to the core scientific methods, technical contents, or any substantive research contributions. The authors led the research, made the scientific decisions, and take full responsibility for the methodology, technical content, experimental results, and final manuscript.

{
    \small
    \bibliographystyle{ieeenat_fullname}
    \bibliography{main}

@String(PAMI  = {IEEE TPAMI})

@String(CVPR  = {CVPR})

@String(ICCV  = {ICCV})

@String(ECCV  = {ECCV})

@String(NIPS  = {NeurIPS})

@String(ICLR  = {ICLR})

@String(IJCAI = {IJCAI})

@String(AAAI  = {AAAI})

@String(WACV  = {WACV})

@String(TMLR  = {TMLR})

@article{huang2023teach,
  author = {Huang, Linjiang and Lu, Kaixin and Song, Guanglu and Wang, Liang and Liu, Si and Liu, Yu and Li, Hongsheng},
  title = {{Teach-DETR}: Better Training {DETR} With Teachers},
  journal = PAMI,
  volume = {45},
  number = {12},
  pages = {15759--15771},
  year = {2023}
}

@inproceedings{geiger2012kitti,
  author = {Geiger, Andreas and Lenz, Philip and Urtasun, Raquel},
  title = {Are We Ready for Autonomous Driving? The {KITTI} Vision Benchmark Suite},
  booktitle = CVPR,
  pages = {3354--3361},
  year = {2012}
}

@inproceedings{shridhar2020alfred,
  author = {Shridhar, Mohit and Thomason, Jesse and Gordon, Daniel and Bisk, Yonatan and Han, Winson and Mottaghi, Roozbeh and Zettlemoyer, Luke and Fox, Dieter},
  title = {{ALFRED}: A Benchmark for Interpreting Grounded Instructions for Everyday Tasks},
  booktitle = CVPR,
  year = {2020}
}

@inproceedings{D-FINE,
  author = {Peng, Yansong and Li, Hebei and Wu, Peixi and Zhang, Yueyi and Sun, Xiaoyan and Wu, Feng},
  booktitle = ICLR,
  pages = {44015--44031},
  title = {{D-FINE}: Redefine Regression Task of {DETRs} as Fine-grained Distribution Refinement},
  year = {2025}
}

@inproceedings{DEIM,
  author = {Huang, Shihua and Lu, Zhichao and Cun, Xiaodong and Yu, Yongjun and Zhou, Xiao and Shen, Xi},
  title = {{DEIM}: {DETR} with Improved Matching for Fast Convergence},
  booktitle = CVPR,
  year = {2025},
  pages = {15162--15171}
}

@article{EdgeCrafter,
  title = {{EdgeCrafter}: Compact {ViTs} for Edge Dense Prediction via Task-Specialized Distillation},
  author = {Longfei Liu and Yongjie Hou and Yang Li and Qirui Wang and Youyang Sha and Yongjun Yu and Yinzhi Wang and Peizhe Ru and Xuanlong Yu and Xi Shen},
  journal = TMLR,
  year = {2026},
  eprint = {2603.18739},
  archivePrefix = {arXiv},
}

@inproceedings{FitNets,
  title = {{FitNets}: Hints for Thin Deep Nets},
  author = {Romero, Adriana and Ballas, Nicolas and Kahou, Samira Ebrahimi and Chassang, Antoine and Gatta, Carlo and Bengio, Yoshua},
  booktitle = ICLR,
  year = {2015}
}

@inproceedings{GIbox,
  title = {General Instance Distillation for Object Detection},
  author = {Dai, Xing and Jiang, Zeren and Wu, Zhao and Bao, Yiping and Wang, Zhicheng and Liu, Si and Zhou, Erjin},
  booktitle = CVPR,
  year = {2021}
}

@inproceedings{RTDETR,
  author = {Zhao, Yian and Lv, Wenyu and Xu, Shangliang and Wei, Jinman and Wang, Guanzhong and Dang, Qingqing and Liu, Yi and Chen, Jie},
  title = {{DETRs} Beat {YOLOs} on Real-time Object Detection},
  booktitle = CVPR,
  year = {2024},
  pages = {16965--16974}
}

@inproceedings{RTDETRv4,
  title = {{RT-DETRv4}: Painlessly Furthering Real-Time Object Detection with Vision Foundation Models},
  author = {Zijun Liao and Yian Zhao and Xin Shan and Yu Yan and Chang Liu and Lei Lu and Xiangyang Ji and Jie Chen},
  booktitle = ECCV,
  year = {2026},
}

@inproceedings{Zagoruyko2017AT,
  title = {Paying More Attention to Attention: Improving the Performance of Convolutional Neural Networks Via Attention Transfer},
  author = {Zagoruyko, Sergey and Komodakis, Nikos},
  booktitle = ICLR,
  year = {2017}
}

@inproceedings{cao2022pkd,
  title = {{PKD}: General distillation framework for object detectors via pearson correlation coefficient},
  author = {Cao, Weihan and Zhang, Yifan and Gao, Jianfei and Cheng, Anda and Cheng, Ke and Cheng, Jian},
  booktitle = NIPS,
  volume = {35},
  pages = {15394--15406},
  year = {2022}
}

@inproceedings{carion2020end,
  title = {End-to-end object detection with transformers},
  author = {Carion, Nicolas and Massa, Francisco and Synnaeve, Gabriel and Usunier, Nicolas and Kirillov, Alexander and Zagoruyko, Sergey},
  booktitle = ECCV,
  pages = {213--229},
  year = {2020},
  organization = {Springer}
}

@inproceedings{chang2023detrdistill,
  title = {{DETRDistill}: A universal knowledge distillation framework for {DETR}-families},
  author = {Chang, Jiahao and Wang, Shuo and Xu, Hai-Ming and Chen, Zehui and Yang, Chenhongyi and Zhao, Feng},
  booktitle = ICCV,
  pages = {6898--6908},
  year = {2023}
}

@inproceedings{chen2017learning,
  title = {Learning Efficient Object Detection Models with Knowledge Distillation},
  author = {Chen, Guobin and Choi, Wongun and Yu, Xiang and Han, Tony and Chandraker, Manmohan},
  booktitle = NIPS,
  year = {2017}
}

@article{chen2024lw,
  title = {{LW-DETR}: A Transformer Replacement to {YOLO} for Real-Time Detection},
  author = {Chen, Qiang and Su, Xiangbo and Zhang, Xinyu and Wang, Jian and Chen, Jiahui and Shen, Yunpeng and Han, Chuchu and Chen, Ziliang and Xu, Weixiang and Li, Fanrong and others},
  journal = {arXiv preprint arXiv:2406.03459},
  year = {2024}
}

@article{deimv2,
  title = {Real-Time Object Detection Meets {DINOv3}},
  author = {Huang, Shihua and Hou, Yongjie and Liu, Longfei and Yu, Xuanlong and Shen, Xi},
  journal = {arXiv preprint arXiv:2509.20787},
  year = {2025}
}

@article{hinton2015distilling,
  title = {Distilling the knowledge in a neural network},
  author = {Hinton, Geoffrey and Vinyals, Oriol and Dean, Jeff},
  journal = {arXiv preprint arXiv:1503.02531},
  year = {2015}
}

@inproceedings{ijcai2024p74,
  title = {{D3ETR}: Decoder Distillation for Detection Transformer},
  author = {Chen, Xiaokang and Chen, Jiahui and Liu, Yan and Tang, Jiaxiang and Zeng, Gang},
  booktitle = IJCAI,
  publisher = {International Joint Conferences on Artificial Intelligence Organization},
  pages = {668--676},
  year = {2024},
}

@inproceedings{lan2026clockdistill,
  title = {{CLoCKDistill}: Consistent location and context aware knowledge distillation for {DETRs}},
  author = {Lan, Qizhen and Tian, Qing},
  booktitle = WACV,
  pages = {7188--7197},
  year = {2026}
}

@inproceedings{lin2014microsoft,
  title = {Microsoft {COCO}: Common objects in context},
  author = {Lin, Tsung-Yi and Maire, Michael and Belongie, Serge and Hays, James and Perona, Pietro and Ramanan, Deva and Doll{\'a}r, Piotr and Zitnick, C Lawrence},
  booktitle = ECCV,
  pages = {740--755},
  year = {2014},
  organization = {Springer}
}

@inproceedings{meng2021conditional,
  title = {Conditional {DETR} for fast training convergence},
  author = {Meng, Depu and Chen, Xiaokang and Fan, Zejia and Zeng, Gang and Li, Houqiang and Yuan, Yuhui and Sun, Lei and Wang, Jingdong},
  booktitle = ICCV,
  pages = {3651--3660},
  year = {2021}
}

@inproceedings{relationKD,
  title = {Relational Knowledge Distillation},
  author = {Park, Wonpyo and Kim, Dongju and Lu, Yan and Cho, Minsu},
  booktitle = CVPR,
  year = {2019}
}

@inproceedings{shao2019objects365,
  title = {Objects365: A Large-Scale, High-Quality Dataset for Object Detection},
  author = {Shao, Shuai and Li, Zeming and Zhang, Tianyuan and Peng, Chao and Yu, Gang and Zhang, Xiangyu and Li, Jing and Sun, Jian},
  booktitle = ICCV,
  pages = {8430--8439},
  year = {2019}
}

@article{simeoni2025dinov3,
  title = {{DINOv3}},
  author = {Sim{\'e}oni, Oriane and Vo, Huy V and Seitzer, Maximilian and Baldassarre, Federico and Oquab, Maxime and Jose, Cijo and Khalidov, Vasil and Szafraniec, Marc and Yi, Seungeun and Ramamonjisoa, Micha{\"e}l and others},
  journal = {arXiv preprint arXiv:2508.10104},
  year = {2025}
}

@inproceedings{wang2019distilling,
  title = {Distilling Object Detectors with Fine-Grained Feature Imitation},
  author = {Wang, Tao and Yuan, Li and Zhang, Xiaopeng and Feng, Jiashi},
  booktitle = CVPR,
  year = {2019}
}

@inproceedings{wang2024kd,
  title = {{KD-DETR}: Knowledge Distillation for Detection Transformer with Consistent Distillation Points Sampling},
  author = {Wang, Yu and Li, Xin and Weng, Shengzhao and Zhang, Gang and Yue, Haixiao and Feng, Haocheng and Han, Junyu and Ding, Errui},
  booktitle = CVPR,
  pages = {16016--16025},
  year = {2024}
}

@inproceedings{wu2024od,
  title = {{OD-DETR}: Online Distillation for Stabilizing Training of Detection Transformer},
  author = {Wu, Shengjian and Sun, Li and Li, Qingli},
  booktitle = IJCAI,
  pages = {1443--1451},
  year = {2024}
}

@inproceedings{wang2024yolov10,
  author = {Wang, Ao and Chen, Hui and Liu, Lihao and Chen, Kai and Lin, Zijia and Han, Jungong and Ding, Guiguang},
  title = {{YOLOv10}: Real-Time End-to-End Object Detection},
  booktitle = NIPS,
  volume = {37},
  year = {2024}
}

@inproceedings{zhang2022dino,
  title = {{DINO}: {DETR} with improved denoising anchor boxes for end-to-end object detection},
  author = {Zhang, Hao and Li, Feng and Liu, Shilong and Zhang, Lei and Su, Hang and Zhu, Jun and Ni, Lionel and Shum, Heung-Yeung},
  booktitle = ICLR,
  year = {2023}
}

@inproceedings{zhu2020deformable,
  title = {Deformable {DETR}: Deformable Transformers for End-to-End Object Detection},
  author = {Zhu, Xizhou and Su, Weijie and Lu, Lewei and Li, Bin and Wang, Xiaogang and Dai, Jifeng},
  booktitle = ICLR,
  year = {2021}
}

@article{yao2021efficient,
  author = {Yao, Zhuyu and Ai, Jiangbo and Li, Boxun and Zhang, Chi},
  title = {Efficient {DETR}: Improving End-to-End Object Detector with Dense Prior},
  journal = {arXiv preprint arXiv:2104.01318},
  year = {2021}
}

@inproceedings{wang2022anchor,
  author = {Wang, Yingming and Zhang, Xiangyu and Yang, Tong and Sun, Jian},
  title = {Anchor {DETR}: Query Design for Transformer-Based Detector},
  booktitle = AAAI,
  pages = {2567--2575},
  year = {2022}
}

@inproceedings{liu2022dab,
  author = {Liu, Shilong and Li, Feng and Zhang, Hao and Yang, Xiao and Qi, Xianbiao and Su, Hang and Zhu, Jun and Zhang, Lei},
  title = {{DAB-DETR}: Dynamic Anchor Boxes Are Better Queries for {DETR}},
  booktitle = ICLR,
  year = {2022}
}

@inproceedings{li2022dn,
  author = {Li, Feng and Zhang, Hao and Liu, Shilong and Guo, Jian and Ni, Lionel M. and Zhang, Lei},
  title = {{DN-DETR}: Accelerate {DETR} Training by Introducing Query {DeNoising}},
  booktitle = CVPR,
  pages = {13619--13627},
  year = {2022}
}

@inproceedings{chen2023group,
  author = {Chen, Qiang and Chen, Xiaokang and Wang, Jian and Zhang, Shan and Yao, Kun and Feng, Haocheng and Han, Junyu and Ding, Errui and Zeng, Gang and Wang, Jingdong},
  title = {Group {DETR}: Fast {DETR} Training with Group-Wise One-to-Many Assignment},
  booktitle = ICCV,
  pages = {6633--6642},
  year = {2023}
}

@inproceedings{guo2021defeat,
  author = {Guo, Jianyuan and Han, Kai and Wang, Yunhe and Wu, Han and Chen, Xinghao and Xu, Chunjing and Xu, Chang},
  title = {Distilling Object Detectors via Decoupled Features},
  booktitle = CVPR,
  pages = {2154--2164},
  year = {2021}
}

@inproceedings{yang2022fgd,
  author = {Yang, Zhendong and Li, Zhe and Jiang, Xiaohu and Gong, Yuan and Yuan, Zehuan and Zhao, Danpei and Yuan, Chun},
  title = {Focal and Global Knowledge Distillation for Detectors},
  booktitle = CVPR,
  pages = {4643--4652},
  year = {2022}
}

@inproceedings{shu2021cwd,
  author = {Shu, Changyong and Liu, Yifan and Gao, Jianfei and Yan, Zheng and Shen, Chunhua},
  title = {Channel-Wise Knowledge Distillation for Dense Prediction},
  booktitle = ICCV,
  pages = {5311--5320},
  year = {2021}
}

@inproceedings{yang2022mgd,
  author = {Yang, Zhendong and Li, Zhe and Shao, Mingqi and Shi, Dachuan and Yuan, Zehuan and Yuan, Chun},
  title = {Masked Generative Distillation},
  booktitle = ECCV,
  pages = {53--69},
  year = {2022}
}

@inproceedings{rezatofighi2019giou,
  author = {Rezatofighi, Hamid and Tsoi, Nathan and Gwak, JunYoung and Sadeghian, Amir and Reid, Ian and Savarese, Silvio},
  title = {Generalized Intersection over Union: A Metric and a Loss for Bounding Box Regression},
  booktitle = CVPR,
  year = {2019}
}

@inproceedings{he2016resnet,
  author = {He, Kaiming and Zhang, Xiangyu and Ren, Shaoqing and Sun, Jian},
  title = {Deep Residual Learning for Image Recognition},
  booktitle = CVPR,
  pages = {770--778},
  year = {2016}
}
}

\end{document}